\documentclass[
]{ceurart}

\usepackage{placeins}
\usepackage{hyperref}

\usepackage{listings}
\newcommand{\hmbert}{\textsc{hmBert}}
\newcommand{\flair}{\textsc{Flair}}
\newcommand{\unk}{\texttt{UNK}\xspace}

\usepackage{todonotes}

\begin{document}

\copyrightyear{2022}
\copyrightclause{Copyright for this paper by its authors.
  Use permitted under Creative Commons License Attribution 4.0
  International (CC BY 4.0).}

\conference{CLEF 2022: Conference and Labs of the Evaluation Forum, 
    September 5--8, 2022, Bologna, Italy}


\title{hmBERT: Historical Multilingual Language Models for Named Entity Recognition}

\author[1]{Stefan Schweter}[%
orcid=0000-0002-7190-2090,
%email=stefan@schweter.it,
%url=https://github.com/stefan-it,
]

\author[2,3]{Luisa März}[
orcid=0000-0003-4542-2437,
%email=luisa.k.maerz@gmail.com,
%url=https://github.com/LuisaMaerz,
]

\author[1]{Katharina Schmid}[
orcid=0000-0001-6057-6640,
]

\author[2]{Erion \c{C}ano}[
orcid=0000-0002-5496-3860,
%email=erion.cano@univie.ac.at,
%url=https://erionc.github.io/,
]

\address[1]{Bayerische Staatsbibliothek München, Digital Library/ Munich Digitization Center, Munich, Germany}

\address[2]{Digital Philology, Research Group Data Mining and Machine Learning, University of Vienna, Austria}

\address[3]{Natural Language Processing Expert Center, Data:Lab, Volkswagen AG, Munich, Germany}

\begin{abstract}
Compared to standard Named Entity Recognition (NER), identifying persons, locations, and organizations in historical texts constitutes a big challenge.
To obtain machine-readable corpora, the historical text is usually scanned and Optical Character Recognition (OCR) needs to be performed. As a result, the historical corpora contain errors. Also, entities like location or organization can change over time, which poses another challenge. Overall, historical texts come with several peculiarities that differ greatly from modern texts and large labeled corpora for training a neural tagger are hardly available for this domain. 
In this work, we tackle NER for historical German, English, French, Swedish, and Finnish by training large historical language models. We circumvent the need for large amounts of labeled data by using unlabeled data for pretraining a language model. We propose \hmbert{}, a historical multilingual BERT-based language model, and release the model in several versions of different sizes. Furthermore, we evaluate the capability of \hmbert{} by solving downstream NER as part of this year's HIPE-2022 shared task and provide detailed analysis and insights. For the Multilingual Classical Commentary coarse-grained NER challenge, our tagger \textit{HISTeria} outperforms the other teams' models for two out of three languages.  
\end{abstract}


\maketitle

\section{Introduction}
Standard Named Entity Recognition (NER) for high resource domains has already been successfully addressed with performances above 90\% F1-score \citep{akbik-etal-2019-pooled,wang2020automated}.
In contrast, NER taggers often fail to achieve satisfying results in the historical domain. 
Since historical datasets usually stem from Optical Character Recognition (OCR) and also include domain shifts, they contain characteristic errors not found in modern text. 
Low-resource fonts like Fraktur pose additional challenges for clean OCR. 
Another problem is that large amounts of labeled data are required when training neural models and only little labeled data exists for historical NER \cite{Ehrmann:221391}.
Because of all these challenges, systems designed for contemporary datasets cannot be applied to the historical domain without adaptations or further training. However, in the last few years, a number of works have shown that it is possible to adapt systems by using different approaches \citep{nerc_hist_survey}.

In this work, we develop a new BERT-based language model \cite{devlin-etal-2019-bert} for the historical context: \hmbert{}.
We tackle NER for historical German, English, French, Swedish, and Finnish.
We use self-supervised learning to pretrain our language model on unlabeled data before we fine-tune the NER tagger on labeled data. This allows to reduce the need for large amounts of labeled training data.
To mitigate the impact of OCR noise in the pretraining corpora, we use a filtering step that allows to control the OCR confidence of the texts.

Another design step in training language models is the choice of the underlying vocabulary. 
\citet{beltagy-etal-2019-scibert} showed that using a domain-specific vocabulary leads to performance improvements compared to using a general domain vocabulary. 
Thus, we use a sub-corpus of our pretraining corpus to create an in-domain vocabulary for the \hmbert{} training.
Finally, we arrive with a powerful \hmbert{} model that establishes state-of-the-art results for three out of four languages on the NewsEye NER dataset~\cite{ahmed_hamdi_2021_4573313}. 
As large language models require a lot of computational resources for pretraining and during inference time, we also provide smaller models. 

Addressing the HIPE-2022 NERC-Coarse task, we also study a single-model vs. one-model approach.
Our comparison shows that fine-tuning \hmbert{} models for each language individually (single-model approach) improves performance compared to models that were fine-tuned on data from all languages (one-model approach). At the same time, however, fine-tuning individual models is much more computationally expensive. The one-model approach is more efficient, achieving similar performance while requiring less computation.

In addition, our final model \textit{HISTeria} is trained by using multi-stage fine-tuning. 
We first fine-tune the multilingual model and evaluate it over the development data of all the different available languages. The resulting hyperparameter configuration is used for another fine-tuning step for each monolingual model. Finally, our detailed study of \hmbert{} also includes experiments with a knowledge-based approach, training an ELECTRA-based language model \cite{Clark2020ELECTRA}, and addressing a tokenization issue. These additional experiments did not enhance performance but represent a suitable starting point for further research. 

Our contributions are i) the comprehensive description of the development of \hmbert{}, ii) the release of \hmbert{} models of different sizes, iii) the release of the \hmbert{} pretraining code, and iv) extensive experiments using \hmbert{} including detailed insights for the community.

This paper is structured as follows: the next section (\ref{sec:hmbert}) describes \hmbert{} and its development in details. We include an explanation of the used datasets, as well as processing, hyperparameter settings, and pretraining steps. We close the section with a downstream task evaluation. Section \ref{sec:hipe} provides insights into the HIPE-2022 Multilingual Commentary Challenge\footnote{\url{https://hipe-eval.github.io/HIPE-2022/}} \cite{hipe2022_condensed_2022}. 
We describe our approach for the shared task submission in detail and provide an analysis of our results. We conclude the paper with Section \ref{sec:conc}.

\section{\hmbert{}: Historical Multilingual BERT Model} \label{sec:hmbert}
\begin{figure}
  \centering
  \includegraphics[width=\linewidth]{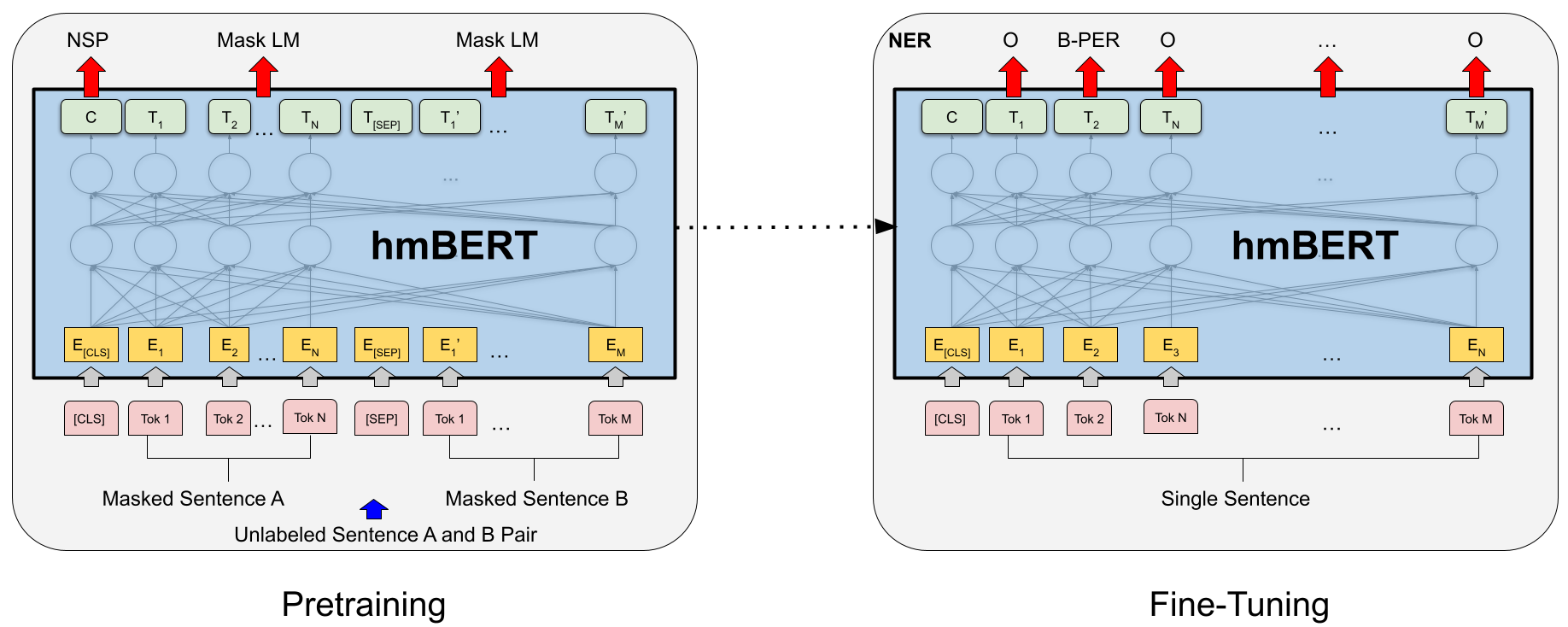}
  \caption{Overall pretraining of our $hmBERT_{32k}$ model and fine-tuning procedure for NER downstream tasks. \texttt{CLS}\xspace is a special symbol added in front of every input example, and \texttt{SEP}\xspace is a special separator token.}
  \label{fig:hmbert-overall}
\end{figure}

In this section, we present \hmbert{} which supports German, English, French, Finnish, and Swedish. We train two different models with different vocabulary sizes: 32,000 and 64,000. We first describe the corpora used for training \hmbert{}, as well as preprocessing and filtering steps to create the pretraining corpus. In addition, we explain the pretraining process and end the section by evaluating the model on a downstream NER task. Figure \ref{fig:hmbert-overall} shows the overall pretraining procedure for \hmbert{} and its application on downstream tasks.

\subsection{Corpora}
For German, French, Swedish and Finnish we use the Europeana newspapers\footnote{\url{http://www.europeana-newspapers.eu/}} provided by the European Library. For English we use a dataset published by the British Library~\cite{bl_corpus}. 
The dataset contains OCR-processed text from digitized books and has also been used by \citet{2021arXiv210511321H} to train historical language models for English.

\subsubsection{Filtering}
OCR full-text for the Europeana newspapers also includes an OCR confidence value. This measure indicates the average OCR confidence for each word of a newspaper\footnote{\url{https://www.clarin.eu/sites/default/files/Nuno_Freire_Europeana_CLARINPLUS.pdf}}.
For German and French we perform a number of characters per year analysis using different (minimum required) OCR confidence thresholds. For German, we test three different thresholds and report the resulting dataset size (see Table \ref{tab:OCR-confidence-german} in the appendix). We use an OCR confidence threshold of 0.60 to get a final dataset of approx. 28\,GB. For French, we test five different OCR confidence values (see Table \ref{tab:OCR-confidence-french} in the appendix) and choose 0.70 so that the resulting dataset size of 27\,GB is comparable to the size of the German dataset. For Finnish and Swedish, we use an OCR confidence threshold of 0.60. However, training data for Swedish and Finnish is very limited. In total, only 1.2\,GB for Finnish, and 1.1\,GB for Swedish are available, thus these corpora are not filtered any further using other OCR confidence thresholds. For English, language filtering using \textsc{langdetect}\footnote{\url{https://github.com/Mimino666/langdetect}} for each book in the corpus is performed. Additionally, we use books published between 1800 and 1900 exclusively. The resulting English corpus has a total size of 24\,GB.

\begin{figure}
  \centering
  \includegraphics[width=\linewidth]{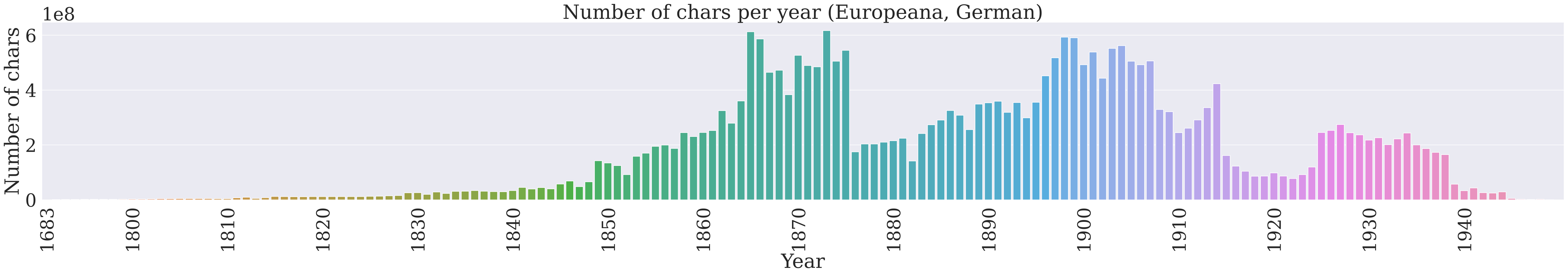}
  \caption{Number of characters per year distribution for filtered German Europeana corpus (1683-1949).}
  \label{fig:german-europeana-stats}
\end{figure}

\begin{figure}
  \centering
  \includegraphics[width=\linewidth]{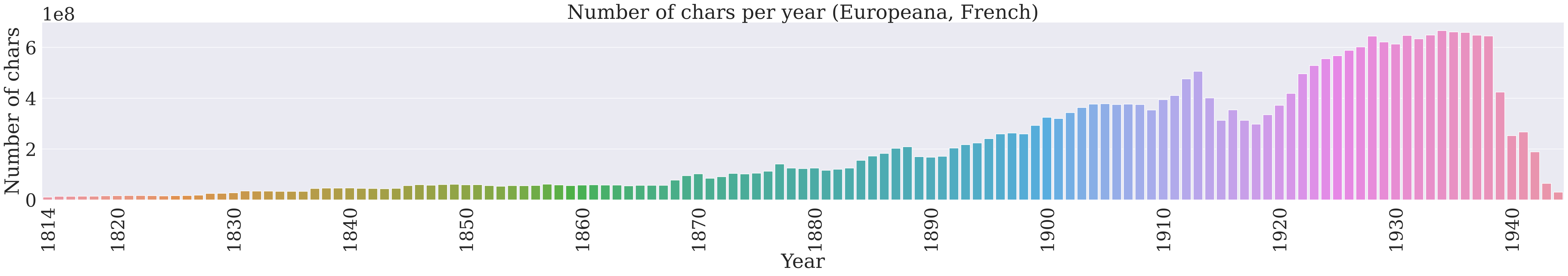}
  \caption{Number of characters per year distribution for filtered French Europeana corpus (1814-1944).}
  \label{fig:french-europeana-stats}
\end{figure}

\begin{figure}
  \centering
  \includegraphics[width=\linewidth]{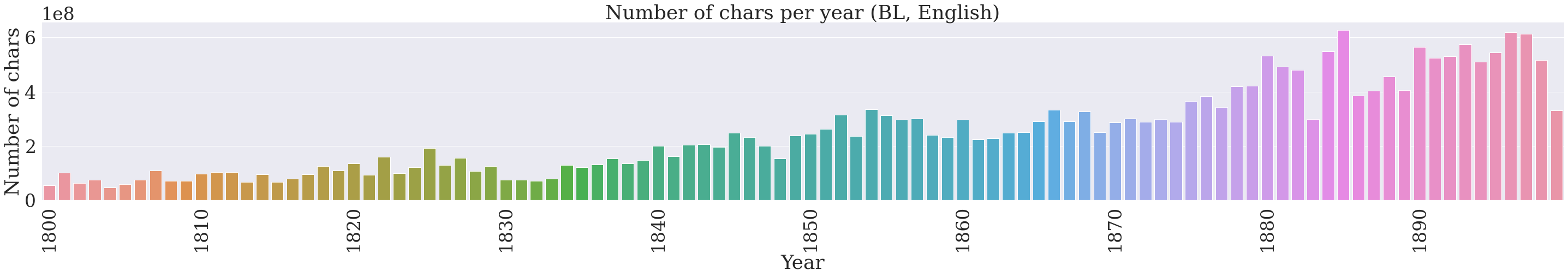}
  \caption{Number of characters per year distribution for filtered English corpus from British Library (1800-1899).}
  \label{fig:bl-stats}
\end{figure}

To get a deeper insight into the filtered corpora, we analyze the distribution of characters over time for each language. Figure \ref{fig:german-europeana-stats} shows the distribution for German. 
The period from 1865 to 1914 is well-covered in the dataset, while the years from 1683 to 1849 and the 20\textsuperscript{th} century are underrepresented. For French, the 20\textsuperscript{th} century is highly covered, but there is only little data available for the 19\textsuperscript{th} century (see Figure \ref{fig:french-europeana-stats}), which contrasts with the German corpus. The English corpus contains texts from the 19\textsuperscript{th} century only and shows good coverage starting from 1850. However, there is only little coverage from 1800 to 1849 (see Figure \ref{fig:bl-stats}). Since both Finnish and Swedish corpora include newspapers from 1900 to 1910 only, we do not analyze the number of characters per year for these datasets.

\subsubsection{Multilingual Vocabulary Generation}
To create a BERT-compatible wordpiece-based vocabulary \cite{DBLP:journals/corr/WuSCLNMKCGMKSJL16}, we use 10GB of each language and train the vocabulary using the Hugging Face Tokenizers library\footnote{\url{https://github.com/huggingface/tokenizers}}. We build a cased vocabulary with no lower casing or accent stripping being performed. 
For Finnish and Swedish we need to upsample\footnote{For upsampling we simply concatenate the original corpus $N$-times to match the desired 10\,GB size per language.} the corpus because both corpora have a size of 1\,GB only.

We create a 32k and 64k vocabulary. Inspired by \citet{rust-etal-2021-good}, we report the subword fertility rate (SFR) and the portion of unknown (\unk{}) tokens per language on various historical NER datasets (see Table \ref{tab:ner-corpora-subword-fertility-rate}). The SFR is defined as the average number of subwords a tokenizer produces per word \cite{rust-etal-2021-good}. 
It indicates how aggressively a tokenizer splits, i.e. whether it over-segments or not. 
As over-segmentation can negatively impact downstream performance, an SFR close to 1 (indicating that the tokenizer vocabulary contains every word in the input text) is optimal. \unk{} tokens are challenging because such tokens are not seen during pretraining and the model cannot provide useful information for them during the fine-tuning phase \cite{pfeiffer-etal-2021-unks}. Table \ref{tab:ner-subword-unks-32k} and Table \ref{tab:ner-subword-unks-64k} show the SFR and portion of \unk{}s in the 32k/64k corpus. French and English have the lowest SFRs, whereas Finnish has the highest rate in both wordpiece-based vocabularies.

% Note: we really used CLEF-HIPE-2020 data from two years ago, because at the time of hmBERT pretraining, HIPE-2022 dataset did not exist!!!
\begin{table*}
  \caption{NER datasets that are used for calculating subword fertility rate and portion of \unk{}s. For English, the development dataset was used due to a missing training split.}
  \label{tab:ner-corpora-subword-fertility-rate}
  \begin{tabular}{ccl}
    \toprule
    Language & NER Corpora\\
    \midrule
    German & CLEF-HIPE-2020~\cite{Ehrmann:281054}, NewsEye~\cite{ahmed_hamdi_2021_4573313} \\
    French & CLEF-HIPE-2020~\cite{Ehrmann:281054}, NewsEye~\cite{ahmed_hamdi_2021_4573313}\\
    English & CLEF-HIPE-2020~\cite{Ehrmann:281054}\\
    Finnish & NewsEye~\cite{ahmed_hamdi_2021_4573313}\\
    Swedish & NewsEye~\cite{ahmed_hamdi_2021_4573313}\\
  \bottomrule
\end{tabular}
\end{table*}

\begin{table*}
  \caption{Subword fertility rate and portion of \unk{}s calculated on NER datasets using a 32k wordpiece-based vocabulary.}
  \label{tab:ner-subword-unks-32k}
  \begin{tabular}{ccl}
    \toprule
    Language & Subword Fertility & \unk{} Portion\\
    \midrule
    German & 1.43 & 0.0004\\
    French & 1.25 & 0.0001\\
    English & 1.25 & 0.0\\
    Finnish & 1.69 & 0.0007\\
    Swedish & 1.43 & 0.0\\
  \bottomrule
\end{tabular}
\end{table*}

\begin{table*}
  \caption{Subword fertility rate and portion of \unk{}s calculated on NER datasets using a 64k wordpiece-based vocabulary.}
  \label{tab:ner-subword-unks-64k}
  \begin{tabular}{ccl}
    \toprule
    Language & Subword Fertility & \unk{} Portion\\
    \midrule
    German & 1.31 & 0.0004\\
    French & 1.16 & 0.0001\\
    English & 1.17 & 0.0\\
    Finnish & 1.54 & 0.0007\\
    Swedish & 1.32 & 0.0\\
  \bottomrule
\end{tabular}
\end{table*}

\subsection{Final Pretraining Corpus}
For common multilingual models such as multilingual BERT \cite[mBERT;][]{devlin-etal-2019-bert}, XLM-RoBERTa~\cite{conneau-etal-2020-unsupervised} or mT5~\cite{xue-etal-2021-mt5} different corpus sampling strategies have been developed to up-/downsample low-/high-resource languages \cite{NEURIPS2019_c04c19c2}. Since our multilingual language model includes five languages only (mBERT covers 104 languages\footnote{\url{https://github.com/google-research/bert/blob/master/multilingual.md}}), we use a similar size for all languages. After upsampling the Swedish and Finnish corpora to 27GB each, we arrive at a total dataset size of 130\,GB. Table \ref{tab:hmbert-final-pretraining-size} shows an overview of the sizes per language included in our final pretraining corpus.
\begin{table*}
  \caption{Size per language of final pretraining corpus for \hmbert{}.}
  \label{tab:hmbert-final-pretraining-size}
  \begin{tabular}{cl}
    \toprule
    Language & Dataset Size\\
    \midrule
    German & 28GB\\
    French & 27GB\\
    English & 24GB\\
    Finnish & 27GB\\
    Swedish & 27GB\\
    \midrule
    Total  & 130GB\\
  \bottomrule
\end{tabular}
\end{table*}
For the \hmbert{} model with a vocabulary size of 32k, we use the official BERT implementation\footnote{\url{https://github.com/google-research/bert\#pre-training-with-bert}} to create pretraining data. Detailed description of all parameters used for the creation of pretraining data can be found in Section \ref{app:final_pretraining_corpus} of the appendix.

\subsection{Models}
We pretrain an \hmbert{} model with a vocabulary size of 32k, further denoted as $hmBERT_{32k}$, and another \hmbert{} model with a vocabulary size of 64k, further denoted as $hmBERT_{64k}$. Inspired by \citet{2022arXiv220313240H}, we also pretrain and release smaller \hmbert{} models, with the number of layers ranging from 2 to 8 and hidden sizes ranging from 128 to 512. Pretraining of the different models is described in detail in Section \ref{app:models_pretraining} of the appendix.

\subsection{Downstream Task Evaluation}
We evaluate the $hmBERT_{32k}$ models on the NewsEye NER dataset~\cite{ahmed_hamdi_2021_4573313}, because this dataset includes most of the languages that \hmbert{} covers (except English), and compare them with the current state-of-the-art reported by \citet{Hamdi2021multilingual}. 
We use the \flair{} \cite{akbik2019flair} library and perform a hyperparameter search (see Table \ref{tab:flair-hyper-param-search-downstream-eval} in appendix) using the common fine-tuning paradigm. Fine-tuning adds a single linear layer to a Transformer and fine-tunes the entire architecture on the NER downstream task. To bridge the difference between subword modeling and token-level predictions, subword pooling is applied to create token-level presentations which are then passed to the final linear layer. 
A common subword pooling strategy is to use the first subtoken to represent the entire token and we also use this strategy in our experiments. 
To train our architecture, we use AdamW \cite{loshchilov2018decoupled} optimizer, a very small learning rate and a fixed number of epochs as a hard-stopping criterion. We evaluate the model performance after each training epoch on the development set and use the best model (strict micro F1-score) for final evaluation. We adopt a one-cycle \cite{2018arXiv180309820S} training strategy, in which the learning rate linearly decreases until it reaches $0$ by the end of the training. 
Tables \ref{tab:newseye-smaller-hmbert-german} - \ref{tab:newseye-smaller-hmbert-swedish} show the performance of our $hmBERT_{32k}$ models compared to the current state-of-the-art. 

\begin{figure}
  \centering
  \includegraphics[width=10cm]{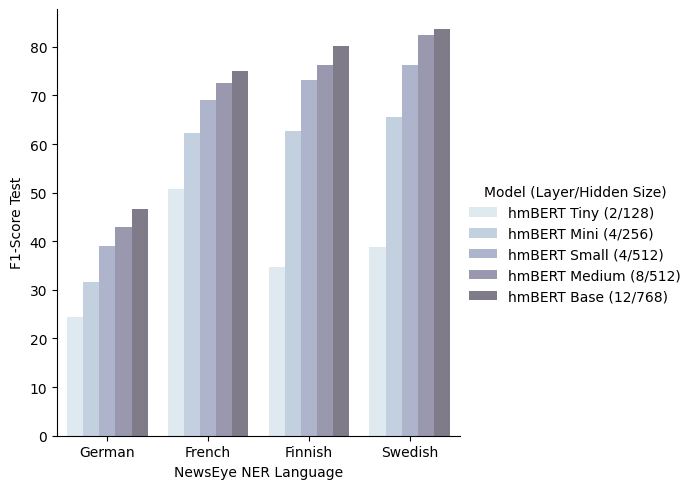}
  \caption{Overview of performance of $hmBERT_{32k}$ smaller models on NewsEye NER datasets. F1-score on the test set is reported here.}
  \label{fig:newseye-evaluation}
\end{figure}

For German, even the $hmBERT_{32k}$ base model could not reach the performance reported by \citet{Hamdi2021multilingual}, that was based on the models developed by \citet{boros_alleviating_2020}. 
The performance difference is 1.64 percentage points. This could be due to the fact that the German NewsEye dataset is very large and the hyperparameter search needed to be extended. 
Furthermore, \citet{Hamdi2021multilingual} proposed a new architecture for handling OCR errors by adding two extra transformer layers, whereas we only performed a standard fine-tuning approach. 
For French our $hmBERT_{32k}$ medium sized model is very close to the result reported by \citet{Hamdi2021multilingual}. 
The $hmBERT_{32k}$ base model outperforms the current best result by +2.7 percentage points. 
The same performance gain can be observed for Finnish and Swedish: The $hmBERT_{32k}$ base model outperforms the current SOTA by 2.41 percentage points for Finnish, and 2.1 percentage points for the Swedish NewsEye dataset. 
Figure \ref{fig:newseye-evaluation} shows an overall performance comparison for the pretrained $hmBERT_{32k}$ smaller models on the NewsEye dataset. 
On average, the performance difference between the 8-layer $hmBERT_{32k}$ medium and the 12-layer $hmBERT_{32k}$ base model is 2.7 percentage points.

\begin{table*}
  \caption{Performance overview of $hmBERT_{32k}$ models on German NewsEye NER dataset.}
  \label{tab:newseye-smaller-hmbert-german}
  \begin{tabular}{lcc}
    \toprule
    Model Name & Development F1-Score & Test F1-Score\\
    \midrule
    $hmBERT_{32k}$ Tiny   & 30.16 & 24.35\\
    $hmBERT_{32k}$ Mini   & 35.74 & 31.54\\
    $hmBERT_{32k}$ Small  & 40.27 & 39.04\\
    $hmBERT_{32k}$ Medium & 43.45 & 43,41\\
    $hmBERT_{32k}$ Base   & 46.17 & 46.66\\
    \midrule
    \citet{Hamdi2021multilingual}         & -     & \textbf{48.3}\\
  \bottomrule
\end{tabular}
\end{table*}

\begin{table*}
  \caption{Performance overview of $hmBERT_{32k}$ models on French NewsEye NER dataset.}
  \label{tab:newseye-smaller-hmbert-french}
  \begin{tabular}{lcc}
    \toprule
    Model Name & Development F1-Score & Test F1-Score\\
    \midrule
    $hmBERT_{32k}$ Tiny   & 60.04 & 50.79\\
    $hmBERT_{32k}$ Mini   & 70.55 & 62.28\\
    $hmBERT_{32k}$ Small  & 75.72 & 69.02\\
    $hmBERT_{32k}$ Medium & 78.99 & 72.51\\
    $hmBERT_{32k}$ Base   & 81.58 & \textbf{75.10}\\
    \midrule
    \citet{Hamdi2021multilingual}         & -     & 72.7\\
  \bottomrule
\end{tabular}
\end{table*}

\begin{table*}
  \caption{Performance overview of $hmBERT_{32k}$ models on Finnish NewsEye NER dataset.}
  \label{tab:newseye-smaller-hmbert-finnish}
  \begin{tabular}{lcc}
    \toprule
    Model Name & Development F1-Score & Test F1-Score\\
    \midrule
    $hmBERT_{32k}$ Tiny   & 30.37 & 34.76\\
    $hmBERT_{32k}$ Mini   & 56.60 & 62.68\\
    $hmBERT_{32k}$ Small  & 64.31 & 73.20\\
    $hmBERT_{32k}$ Medium & 69.95 & 76.34\\
    $hmBERT_{32k}$ Base   & 76.05 & \textbf{80.11}\\
    \midrule
    \citet{Hamdi2021multilingual}         & -     & 77.7\\
  \bottomrule
\end{tabular}
\end{table*}

\begin{table*}
  \caption{Performance overview of $hmBERT_{32k}$ models on Swedish NewsEye NER dataset.}
  \label{tab:newseye-smaller-hmbert-swedish}
  \begin{tabular}{lcc}
    \toprule
    Model Name & Development F1-Score & Test F1-Score\\
    \midrule
    $hmBERT_{32k}$ Tiny   & 43.65 & 38.91\\
    $hmBERT_{32k}$ Mini   & 64.05 & 65.58\\
    $hmBERT_{32k}$ Small  & 73.47 & 76.29\\
    $hmBERT_{32k}$ Medium & 78.07 & 82.47\\
    $hmBERT_{32k}$ Base   & 81.13 & \textbf{83.60}\\
    \midrule
    \citet{Hamdi2021multilingual}         & -     & 81.5\\
  \bottomrule
\end{tabular}
\end{table*}

\section{HIPE-2022: Multilingual Classical Commentary Challenge} \label{sec:hipe}
We participated in the Multilingual Classical Commentary Challenge (MCC) that was newly introduced in the 2022 edition of HIPE \cite{ehrmann_introducing_2022} with our tagger being denoted as \textit{HISTeria}. 
The challenge requires participants to work with historical classical commentaries in at least two different languages and to develop solutions for Named Entity Recognition, Classification, and/or Linking. 
\textit{HISTeria} aims to detect and classify named entities according to coarse-grained types (NERC-Coarse task) and is described in more detail in this section. 

\subsection{Data}
A classical commentary is a scholarly publication that aims to facilitate the reading and understanding of classical works of literature by providing additional information such as translations or bibliographic references. Apart from the challenges that are common to historical texts, commentaries have other characteristics that may complicate Named Entity Recognition and Classification: they frequently cite the original literary text, making them inherently multilingual, and they often use abbreviations to convey information more concisely. For the Multilingual Classical Commentary Challenge, HIPE\footnote{\url{https://github.com/hipe-eval/HIPE-2022-data/blob/main/documentation/README-ajmc.md}} has chosen a single dataset that was created in the context of the Ajax MultiCommentary project\footnote{\url{https://mromanello.github.io/ajax-multi-commentary/}} (\texttt{ajmc} dataset). The dataset contains excerpts from commentaries published in the 19\textsuperscript{th} century in English, French, and German. The French texts date from 1886, the German ones from 1853 and 1894, and the English ones from 1881 and 1896. This emphasis on the second half of the 19\textsuperscript{th} century fits well with the temporal distribution of our pretraining data for English and German. Apart from standard entity types like \textit{person} or \textit{location}, the dataset also includes domain-specific annotations like the \textit{scope} and \textit{work} entity type for bibliographic references. Additional dataset statistics can be found in Table \ref{tab:ajmc-stats} and in the HIPE 2022 Overview paper \cite{hipe2022_condensed_2022}.

\begin{table*}
  \caption{Dataset statistics about \texttt{ajmc} dataset.}
  \label{tab:ajmc-stats}
  \begin{tabular}{lrr}
    \toprule
    Language & Training Sentences & Development Sentences\\
    \midrule
    German  & 1,024 & 192\\
    English & 1,154 & 252\\
    French  &   894 & 202\\
  \bottomrule
\end{tabular}
\end{table*}

\subsection{Single-Models vs. One-Model Approach}
In preliminary experiments models that are independently fine-tuned for each language (single-model approach) and a model that uses training data from all languages (one-model approach) are compared. We perform hyperparameter searches for the two approaches. The relevant hyperparameters for fine-tuning models are shown in the appendix (Table \ref{tab:flair-hyper-param-search-final-first}). We use the \flair{} library for all experiments. For the one-model approach, a breakdown analysis for each language is performed after determining the best hyperparameter configuration. This is compared to the three independently fine-tuned models for each language. For German, the one-model approach is +0.47 percentage points better than the single-model approach. For English, the one-model approach performs slightly worse (-0.13 percentage points) and for French, the single-model approach outperforms the one-model by 0.6 percentage points. However, the single-model approach requires fine-tuning of 120 models, whereas the one-model approach only needs 40 models to be fine-tuned for hyperparameter search. To save resources, we decided to use the one-model approach for further experiments. The performance comparison on the \texttt{ajmc} dataset is shown in Table \ref{tab:ajmc-preliminary-results}.

\begin{table*}
  \caption{Performance comparison for NERC-coarse between single-model and one-model approach on \texttt{ajmc} development dataset. Numbers express F1-score calculated by using the strict evaluation regime.}
  \label{tab:ajmc-preliminary-results}
  \begin{tabular}{lcc}
    \toprule
    Language & Single-Model & One-Model\\
    \midrule
    German  & 86.21 & \textbf{86.68}\\
    English & \textbf{84.98} & 84.85\\
    French  & \textbf{85.69} & 85.09\\
  \bottomrule
\end{tabular}
\end{table*}

\subsection{Multi-Stage Fine-Tuning} 
\citet{wang2022damonlp} proposed a knowledge-based system for multilingual NER using a multi-stage fine-tuning approach for the MultiCoNER SemEval 2022 task\footnote{\url{https://multiconer.github.io/}}. The first stage of multi-stage fine-tuning refers to training a multilingual model on data from different languages. In the second stage, this fine-tuned multilingual model is used as a starting point for training a monolingual model. We adapt this approach for our final system: in the first stage, we fine-tune one multilingual model over the training data of all three languages (German, English, and French) and optimize over all development data (one-model approach) using a hyperparameter search. We select the best hyperparameter configuration as a combination of batch size, the number of epochs, and the learning rate, which results in five models (because of five different random seeds). The hyperparameter search grid for the different stages is shown in Section \ref{app:multistage_finetuning} in the appendix. From these five models, we choose the one with the highest F1-score on the development set for second stage fine-tunings. In the second stage, we use the best model from the first stage and fine-tune single models for each language with a hyperparameter search on the development set. For each language, we select the best hyperparameter configuration and choose the best performing model with the highest F1-score on the development set. In preliminary experiments, this multi-stage fine-tuning approach boosts performance by 1.23 percentage points on average compared to results in the first stage.

For our final submission, $hmBERT_{32k}$ achieves the best results during the first-stage of fine-tuning with a batch size of $4$, $10$ fine-tuning epochs and a learning rate of $5e-05$. This results in an average F1-score of 86.89 on the (combined) development sets for \texttt{ajmc}. The best hyperparameter configuration for $hmBERT_{64k}$ can be achieved when using a batch size of $8$, $10$ epochs of fine-tuning and a learning rate of $3e-05$. This results in an overall F1-score of 86.69 percentage points. Thus, $hmBERT_{64k}$ is slightly worse than $hmBERT_{32k}$ (-0.2 percentage points). Table \ref{tab:final-dev-results} shows the performance for our final submissions using $hmBERT_{32k}$ and $hmBERT_{64k}$ for all languages in the \texttt{ajmc} dataset. We report strict and fuzzy F1-scores using the official HIPE-scorer\footnote{\url{https://github.com/hipe-eval/HIPE-scorer}}. We exclude document-level scores for better readability.

\begin{table*}
  \caption{Final results on \texttt{ajmc} development dataset for all languages using best models after multi-stage fine-tuning. Results are reported with official HIPE scorer.}
  \label{tab:final-dev-results}
  \begin{tabular}{lccc}
    \toprule
    Submission ID & Hyperparameter Configuration & Strict F1-Score & Fuzzy F1-Score\\
    \midrule
    German ($hmBERT_{32k}$) - 1  & bs8-e05-lr3e-05  & 91.5 & \textbf{94.2}\\
    German ($hmBERT_{64k}$) - 2  & bs8-e10-lr3e-05 & \textbf{92.0} & 93.9\\
    \midrule
    English ($hmBERT_{32k}$) - 1 & bs4-e10-lr3e-05 & \textbf{89.1} & 92.9\\
    English ($hmBERT_{64k}$) - 2 & bs8-e10-lr3e-05 & 88.0 & \textbf{93.8}\\
    \midrule
    French ($hmBERT_{32k}$) - 1  & bs4-e10-lr3e-05 & \textbf{86.8} & \textbf{93.1}\\
    French ($hmBERT_{64k}$) - 2  & bs4-e10-lr5e-05 & 85.9 & 93.0\\
  \bottomrule
\end{tabular}
\end{table*}

\subsection{\textit{HISTeria} Results}
Table \ref{tab:final-test-results} shows an overview of \textit{HISTeria} compared to the runs of other teams in the HIPE-2022 shared task\footnote{\url{https://github.com/hipe-eval/HIPE-2022-eval/}} .
% ToDo: check if for camera-ready version again.
%\todo{Someone should double-check it: 
%Notice: because we ranked it on Strict F1-score, the corresponding rank is not identical to Fuzzy F1-score rank! Please use system id in the HIPE-2022-eval repo for corresponding Fuzzy F1-score.}
%https://github.com/hipe-eval/HIPE-2022-eval/blob/main/HIPE_2022_evaluation_results.md#dataset-ajmc
%
\begin{table*}
  \caption{Final results on \texttt{ajmc} test dataset for all languages compared to other participants in the HIPE-2022 shared task. HISTeria denotes our system. Rank is ordered by strict F1-score.}
  \label{tab:final-test-results}
  \begin{tabular}{lllcc}
    \toprule
    Rank & Language & Submission ID             & Strict F1-Score & Fuzzy F1-Score\\
    \midrule
    1    & German   & L3i (team 2) - 2          & 93.4            & 95.2\\
    2    & German   & HISTeria ($hmBERT_{32k}$) - 1 & 91.3            & 93.7\\
    3    & German   & HISTeria ($hmBERT_{64k}$) - 2 & 91.2            & 94.5\\
    4    & German   & L3i (team 2) - 1          & 90.8            & 93.4\\
    5    & German   & Neural baseline                  & 81.8            & 87.3\\
    \midrule
    1    & English  & HISTeria ($hmBERT_{64k}$) - 2 & 85.4            & 91.0\\
    2    & English  & L3i (team 2) - 1          & 85.0            & 89.4\\
    3    & English  & L3i (team 2) - 2          & 84.1            & 88.4\\
    4    & English  & HISTeria ($hmBERT_{32k}$) - 1 & 81.9            & 89.9\\
    5    & English  & Neural baseline                  & 73.6            & 82.8\\
    \midrule
    1    & French   & HISTeria ($hmBERT_{64k}$) - 2 & 84.2            & 88.0\\
    2    & French   & HISTeria ($hmBERT_{32k}$) - 1 & 83.3            & 88.8\\
    3    & French   & L3i (team 2) - 2          & 82.6            & 87.2\\
    4    & French   & L3i (team 2) - 1          & 79.8            & 86.0\\
    5    & French   & Neural baseline                  & 74.1            & 82.5\\
  \bottomrule
\end{tabular}
\end{table*}

To gain a better understanding of our models, we use the attribute-aided evaluation proposed by \citet{fu2020interpretable}. In order to highlight the strengths and weaknesses of different models, they analyze how model performance varies with regard to certain attributes. In the case of NER, properties that may influence performance are i) how consistently a given surface form of a token or an entity is labelled across a dataset (\texttt{tCon} and \texttt{eCon}), ii) how often a given token or entity appears in the dataset (\texttt{tFre} and \texttt{eFre}), iii) the number of tokens that make up an entity (\texttt{eLen}) or sentence (\texttt{sLen}) as well as iv) the relative number of out-of-vocabulary words and entities per sentence (\texttt{oDen} and \texttt{eDen}). Using the implementation by \citet{fu2020interpretable}, we distribute the values into buckets and compute the strict F1-score for each bucket. Table \ref{tab:corr-std-attr} shows Spearman's rank correlation coefficient as a measure of how well the attribute correlates with the F1-score, and the standard deviation of the F1-score to indicate how strongly the attribute influences performance. We omit results that are not statistically significant.

For the two German models, none of the attributes seem to correlate with performance in a statistically significant way. For the English and French models, performance correlates directly and positively with the consistency of the token labels. The standard deviation of 10\% (French) and 8-9\% (English) of the F1-score indicates that this attribute has a marked impact on performance. For French $hmBERT_{32k}$, entity length influences performance to the same degree. In this case, performance gets worse the more tokens an entity has. The impact of entity length on English $hmBERT_{32k}$ and $hmBERT_{64k}$ is less strong but still notable (standard deviation of 6\% and 1\% respectively). In addition to entity length, the amount of words that did not feature in the training set also correlates negatively with the performance of English $hmBERT_{32k}$.

\begin{table*}
  \caption{Spearman's rank correlation coefficient and standard deviation of models' F1-score depending on different attribute values. We omit results that are not statistically significant.}
  \label{tab:corr-std-attr}
  \begin{tabular}{llcc}
    \toprule
    Model                   & Attribute & Spearman & Standard Deviation \\
    \midrule
    English $hmBERT_{32k}$  & \texttt{tCon}      & 1.0       & 0.09 \\
                            & \texttt{eLen}      & -1.0      & 0.06 \\
                            & \texttt{oDen}      & -1.0      & 0.09 \\
    \midrule
    English $hmBERT_{64k}$  & \texttt{tCon}      & 1.0       & 0.08 \\
                            & \texttt{eLen}      & -1.0      & 0.01 \\
    \midrule
    French $hmBERT_{32k}$   & \texttt{tCon}      & 1.0       & 0.10 \\
                            & \texttt{eLen}      & -1.0      & 0.10 \\
    \midrule
    French $hmBERT_{64k}$   & \texttt{tCon}      & 1.0       & 0.10 \\
  \bottomrule
\end{tabular}
\end{table*}

\subsection{Challenges}
We also experimented with the knowledge-based system for multilingual NER that was proposed by \citet{wang2022damonlp}. We used their implementation to enrich the original \texttt{ajmc} datasets with a knowledge base and implemented their context approach in the \flair{} library. More precisely, we used the FLERT approach \cite{schweter2020flert} and utilized the knowledge-base enriched context as the left context for each training example. A left context size of 128 performs best in the experiments. However, the final result was slightly worse than using no context at all. This may be due to the fact that a contemporary, general-purpose knowledge base (Wikipedia) was used. A domain-specific knowledge base may yield better results. As the preliminary results were slightly worse than our main baseline, we did not conduct further experiments with this knowledge-based system.

We calculated the portion of \unk{}s in the German \texttt{ajmc} dataset and found that the portion rate of 16.3\,\% is unreasonably high. We discovered that the German \texttt{ajmc} dataset contains long-s characters, unlike the Europeana Newspaper corpora which were used to train a vocabulary. As a consequence, the \hmbert{} tokenizer is not able to handle tokens that include long-s characters, resulting in \unk{}s. For our final system, we manually replaced all long-s characters with a normal s character to circumvent the \unk{} problem. In upcoming versions of our \hmbert{} models, we will add this replacement step in the tokenizer configuration directly. Furthermore, we also trained an ELECTRA model \cite{Clark2020ELECTRA} for 1M steps on the same pretraining corpus as the $hmBERT_{32k}$ model. We found that the downstream performance on NewsEye datasets was 1 to 3 percentage points worse than $hmBERT_{32k}$ and -0.28 percentage points worse on the \texttt{ajmc} dataset. We have therefore decided not to release the model yet. 

\subsection{Community Contributions}
To foster research on language and NER models for the historical domain, we publicly release our pretrained and fine-tuned models on the Hugging Face Model Hub\footnote{\url{https://huggingface.co/}} under the \texttt{dbmdz} namespace\footnote{\url{https://huggingface.co/dbmdz}}. We also publicly release all code that was used for fine-tuning models\footnote{\url{https://github.com/dbmdz/clef-hipe}}. Table \ref{tab:community-contributions} shows an overview of released models for our HIPE-2022 submission, including the model identifier on the Hugging Face Model Hub. All models are released under a permissive MIT license. Additionally, we added dataset support for all HIPE-2022 NER datasets into \flair{} library\footnote{Added in \flair{} version $0.11$: \url{https://github.com/flairNLP/flair/releases/tag/v0.11}}.

\begin{table*}
  \caption{Community contributions for our HIPE-2022 submission: Pretrained language models and fine-tuned NER models are publicly available on the Hugging Face Model Hub.}
  \label{tab:community-contributions}
  \scriptsize
  \begin{tabular}{ll}
    \toprule
    Model Description & Model Name\\
    \midrule
    $hmBERT_{32k}$ Tiny Model   & \href{https://huggingface.co/dbmdz/bert-tiny-historic-multilingual-cased}
    {\texttt{dbmdz/bert-tiny-historic-multilingual-cased}}\\
    $hmBERT_{32k}$ Mini Model   & \href{https://huggingface.co/dbmdz/bert-mini-historic-multilingual-cased}
    {\texttt{dbmdz/bert-mini-historic-multilingual-cased}}\\
    $hmBERT_{32k}$ Small Model  & \href{https://huggingface.co/dbmdz/bert-small-historic-multilingual-cased}
    {\texttt{dbmdz/bert-small-historic-multilingual-cased}}\\
    $hmBERT_{32k}$ Medium Model & \href{https://huggingface.co/dbmdz/bert-medium-historic-multilingual-cased}
    {\texttt{dbmdz/bert-medium-historic-multilingual-cased}}\\
    $hmBERT_{32k}$ Base Model   & \href{https://huggingface.co/dbmdz/bert-base-historic-multilingual-cased}
    {\texttt{dbmdz/bert-base-historic-multilingual-cased}}\\
    $hmBERT_{64k}$ Base Model   & \href{https://huggingface.co/dbmdz/bert-base-historic-multilingual-64k-td-cased}
    {\texttt{dbmdz/bert-base-historic-multilingual-64k-td-cased}}\\
    \midrule
    NER First Stage ($hmBERT_{32k}$) & \href{https://huggingface.co/dbmdz/flair-hipe-2022-ajmc-all}{\texttt{dbmdz/flair-hipe-2022-ajmc-all}}\\
    NER First Stage ($hmBERT_{64k}$) & \href{https://huggingface.co/dbmdz/flair-hipe-2022-ajmc-all-64k}{\texttt{dbmdz/flair-hipe-2022-ajmc-all-64k}}\\
    \midrule
    NER Second Stage - German ($hmBERT_{32k}$) & \href{https://huggingface.co/dbmdz/flair-hipe-2022-ajmc-de}{\texttt{dbmdz/flair-hipe-2022-ajmc-de}}\\
    NER Second Stage - English ($hmBERT_{32k}$) & \href{https://huggingface.co/dbmdz/flair-hipe-2022-ajmc-en}{\texttt{dbmdz/flair-hipe-2022-ajmc-en}}\\
    NER Second Stage - French ($hmBERT_{32k}$) & \href{https://huggingface.co/dbmdz/flair-hipe-2022-ajmc-fr}{\texttt{dbmdz/flair-hipe-2022-ajmc-fr}}\\
    \midrule
    NER Second Stage - German ($hmBERT_{64k}$) & \href{https://huggingface.co/dbmdz/flair-hipe-2022-ajmc-de-64k}{\texttt{dbmdz/flair-hipe-2022-ajmc-de-64k}}\\
    NER Second Stage - English ($hmBERT_{64k}$) & \href{https://huggingface.co/dbmdz/flair-hipe-2022-ajmc-en-64k}{\texttt{dbmdz/flair-hipe-2022-ajmc-en-64k}}\\
    NER Second Stage - French ($hmBERT_{64k}$) & \href{https://huggingface.co/dbmdz/flair-hipe-2022-ajmc-fr-64k}{\texttt{dbmdz/flair-hipe-2022-ajmc-fr-64k}}\\
  \bottomrule
\end{tabular}
\end{table*}

\section{Conclusion} \label{sec:conc}
We presented \hmbert{}, a new multilingual BERT-based language model for historical data. \hmbert{} is composed of German, French, English, Finnish, and Swedish unsupervised corpora of historical OCR-processed texts. The corpora have been filtered for OCR confidence as well as sampled so that each language contributes a similar amount of data to the model. The underlying vocabulary is also derived from each of the languages used for \hmbert{}. In our temporal analysis of the pretraining corpora, we have found that data from the 18\textsuperscript{th} and 19\textsuperscript{th} century is unevenly distributed across the different languages. For future models, we are looking for additional datasets to balance this representation. We evaluated two \hmbert{} models of different sizes with downstream Named Entity Recognition. For the NewsEye dataset \hmbert{} established a new state-of-the-art for three out of four languages: French, Finnish, and Swedish. For the 2022 HIPE Multilingual Classical Commentary Challenge, our \textit{HISTeria} system could outperform the other systems for two out of three languages. Using multi-stage fine-tuning together with the multilingual BERT-based model led the model to its optimal performance. Detailed analysis showed the benefits of all of \hmbert{}s design choices, as well as interesting findings for future research. Our contributions include all of the trained \hmbert{} models and our source code, which are made publicly available.  

\begin{acknowledgments}
  We would like to thank Google's TPU Research Cloud (TRC) program for giving us access to TPUs that were used for training our \hmbert{} models.
  We would also like to thank Hugging Face for providing the ability to host and perform inferencing of our models on the Hugging Face Model Hub.
  
\end{acknowledgments}

%%
%% Define the bibliography file to be used
\bibliography{sample-ceur}

\clearpage

\appendix

\section{hmBERT: Historical Multilingual BERT Model}

\subsection{Corpora Filtering}

\FloatBarrier
\begin{table*}[ht!]
  \caption{Word level OCR confidence thresholds for German. Bold OCR confidence is used for the final corpus.}
  \label{tab:OCR-confidence-german}
  \begin{tabular}{ccl}
    \toprule
    OCR Confidence & Dataset Size\\
    \midrule
    \textbf{0.60} & 28GB \\
    0.65 & 18GB \\
    0.70 & 13GB \\
  \bottomrule
\end{tabular}
\end{table*}

\begin{table*}[ht!]
  \caption{Word level OCR confidence thresholds for French. Bold OCR confidence is used for the final corpus.}
  \label{tab:OCR-confidence-french}
  \begin{tabular}{ccl}
    \toprule
    OCR Confidence & Dataset Size\\
    \midrule
    0.60 & 31GB \\
    0.65 & 27GB \\
    \textbf{0.70} & 27GB \\
    0.75 & 23GB \\
    0.80 & 11GB \\
  \bottomrule
\end{tabular}
\end{table*}

\FloatBarrier

\subsection{Final Pretraining Corpus}\label{app:final_pretraining_corpus}
\FloatBarrier

For creation of the pretraining data, we use the same parameters as BERTurk~\cite{stefan_schweter_2020_3770924}: maximum sequence length = 512, maximum predictions per sequence = 75, masked language probability rate = 0.15, duplication factor = 5. Due to hardware limitations, we split the pretraining corpus into chunks of 1GB and create pretraining data for each chunk individually. For the \hmbert{} model with a vocabulary size of 64k we use the official implementation\footnote{\url{https://github.com/tensorflow/models/blob/27fb855b027ead16d2616dcb59c67409a2176b7f/official/legacy/bert/README.md\#pre-training}} with the same parameters as for the 32k model, but we increase the maximum predictions per sequence to 76.

\subsection{Models}\label{app:models_pretraining}
\FloatBarrier

\begin{figure}[ht!]
  \centering
  \includegraphics[width=8cm]{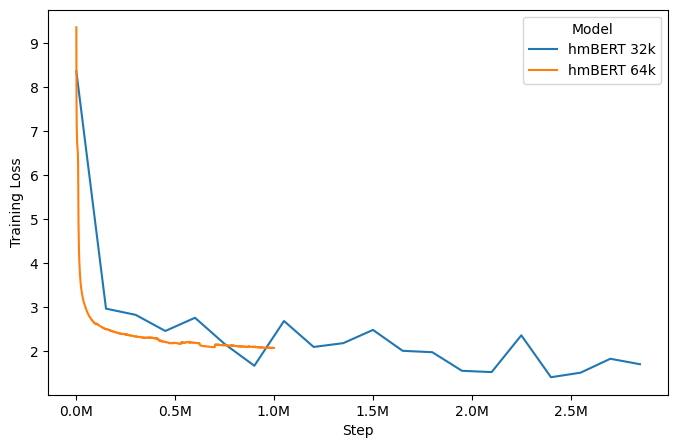}
  \caption{Overview of pretraining loss for $hmBERT_{32k}$ and $hmBERT_{64k}$.}
  \label{fig:hmbert_training_losses}
\end{figure}

We use the official BERT implementation\footnote{\url{https://github.com/google-research/bert}} for pretraining $hmBERT_{32k}$. $hmBERT_{64k}$ is trained with the recently proposed ``token dropping'' approach by \citet{hou-etal-2022-token}. Using this approach, unimportant tokens starting from an intermediate layer in the model are dropped to make the model focus on important tokens more efficiently, which makes model pretraining faster compared to the original BERT implementation. For both pretraining approaches, we use a maximum sequence length of 512 for the full training time. For the pretraining of $hmBERT_{32k}$ a batch size of 128 is used for 3M training steps. Pretraining was done on a v3-32 TPU pod within 67 hours. The pretraining of $hmBERT_{64k}$ was done on a single v4-8 TPU with a batch size of 512 for 1M steps within 114 hours. Figure \ref{fig:hmbert_training_losses} shows the pretraining loss for $hmBERT_{32k}$ and $hmBERT_{64k}$. The final $hmBERT_{32k}$ has 110.62M, whereas $hmBERT_{64k}$ has 135.19M parameters due to the increased vocabulary size.

% New paragraph
For better comparability, we measure the number of total subtokens seen during pretraining\footnote{Total number of subtokens during pretraining can be calculated as multiplication of training steps, batch size and sequence length} and the number of total subtokens of the pretraining corpus for our two \hmbert{} models. More precisely, $hmBERT_{32k}$ has seen 196B subtokens during pretraining, whereas the pretraining corpus has a total size of 42B subtokens. This results in 4.7 pretraining epochs over the corpus. Our $hmBERT_{64k}$ model has seen 262B subtokens during pretraining. Because of the larger vocabulary size, the number of subtokens for the corpus is 39B. This results in 6.7 pretraining epochs over the corpus.

\begin{figure}[ht!]
  \centering
  \includegraphics[width=8cm]{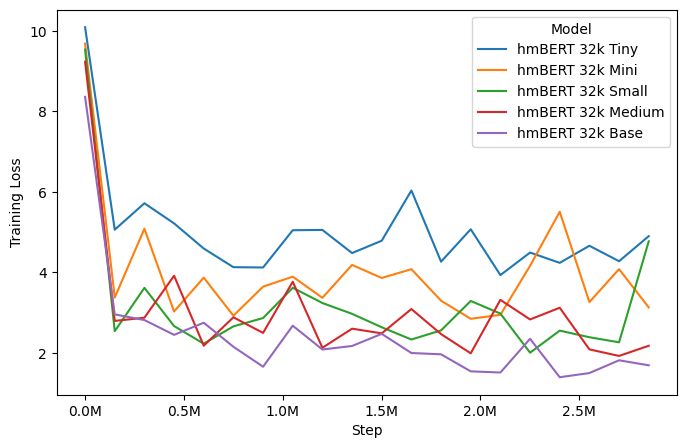}
  \caption{Overview of pretraining loss for smaller $hmBERT_{32k}$ models.}
  \label{fig:hmbert_smaller_training_losses}
\end{figure}

For the smaller models, we use the same pretraining data and hyperparameter as for the base $hmBERT_{32k}$ model and pretrain them on a v3-32 TPU pod. Table \ref{tab:smaller-hmbert-models} shows an overview of pretrained models, including their model size, number of parameters and pretraining time. Figure \ref{fig:hmbert_smaller_training_losses} shows an overview of pretraining loss for all smaller $hmBERT_{32k}$ models.

\begin{table*}[ht!]
  \caption{Overview of smaller $hmBERT_{32k}$ models with their corresponding model size, number of parameters and pretraining time.}
  \label{tab:smaller-hmbert-models}
  \begin{tabular}{lccrr}
    \toprule
    Model Name & Number of Layers & Hidden Size & Parameters & Pretraining Time\\
    \midrule
    $hmBERT_{32k}$ Tiny & 2 & 128 & 4.58M & 4.3s / 1k steps\\
    $hmBERT_{32k}$ Mini & 4 & 256 & 11.55M & 10.5s / 1k steps\\
    $hmBERT_{32k}$ Small & 4 & 512 & 29.52M & 20.7s / 1k steps\\
    $hmBERT_{32k}$ Medium & 8 & 512 & 42.13M & 35.0s / 1k steps\\
    \midrule
    $hmBERT_{32k}$ Base & 12 & 768 & 110.62M & 80.0s / 1k steps\\
  \bottomrule
\end{tabular}
\end{table*}

\FloatBarrier

\subsection{Downstream Task Evaluation}

\FloatBarrier

\begin{table*}[ht!]
  \caption{Hyperparameter search for downstream evaluation on NewsEye NER dataset.}
  \label{tab:flair-hyper-param-search-downstream-eval}
  \begin{tabular}{ll}
    \toprule
    Parameter & Values\\
    \midrule
    Batch Size & $[4, 8]$\\
    Epoch & $[5, 10]$\\
    Learning Rate & $[3e-05, 5e-05]$\\
    Seed & $[1, 2 , 4 , 5]$\\
  \bottomrule
\end{tabular}
\end{table*}

\FloatBarrier

\section{HIPE-2022: Multilingual Classical Commentary Challenge}

\subsection{Multi-Stage Fine-Tuning}\label{app:multistage_finetuning}

\FloatBarrier

\begin{table*}[ht!]
  \caption{Hyperparameter search during the first stage of NER model fine-tuning.}
  \label{tab:flair-hyper-param-search-final-first}
  \begin{tabular}{ll}
    \toprule
    Parameter & Values\\
    \midrule
    Batch Size & $[4, 8, 16]$\\
    Epoch & $[10]$\\
    Learning Rate & $[1e-05, 2e-05, 3e-05, 4e-05, 5e-05]$\\
    Seed & $[1, 2 , 4 , 5]$\\
  \bottomrule
\end{tabular}
\end{table*}

\begin{table*}[ht!]
  \caption{Hyperparameter search during the second stage of NER model fine-tuning.}
  \label{tab:flair-hyper-param-search-final-second}
  \begin{tabular}{ll}
    \toprule
    Parameter & Values\\
    \midrule
    Batch Size & $[4, 8]$\\
    Epoch & $[5, 10]$\\
    Learning Rate & $[3e-05, 5e-05]$\\
    Seed & $[1, 2 , 4 , 5]$\\
  \bottomrule
\end{tabular}
\end{table*}

\FloatBarrier

As a batch size of 16 and learning rates of $1e-05$ and $2e-05$ do not perform well, we exclude them when performing hyperparameter search with $hmBERT_{64k}$. 

\end{document}